%% file: main.tex
\documentclass[sigconf,nonacm]{acmart}
\setcopyright{none}
\renewcommand\footnotetextcopyrightpermission[1]{}

\input{preamble}

\title{LentiAvatar: Pseudo-Multiview Reconstruction and Subpixel Prism Rendering for Real-Time Stereoscopic Communication}

\author{Chufeng Fang}
\email{fangchf3@mail2.sysu.edu.cn}
\affiliation{%
  \institution{Sun Yat-sen University}
  \city{Guangzhou}
  \country{China}
}

\author{Dongdong Teng}
\email{tengdd@mail.sysu.edu.cn}
\affiliation{%
  \institution{Sun Yat-sen University}
  \city{Guangzhou}
  \country{China}
}

\author{Lilin Liu}
\authornote{Corresponding author.}
\email{liullin@mail.sysu.edu.cn}
\affiliation{%
  \institution{Sun Yat-sen University}
  \city{Guangzhou}
  \country{China}
}

\begin{document}

\input{sec/0_abstract}



\begin{teaserfigure}
\centering
\includegraphics[width=0.98\textwidth]{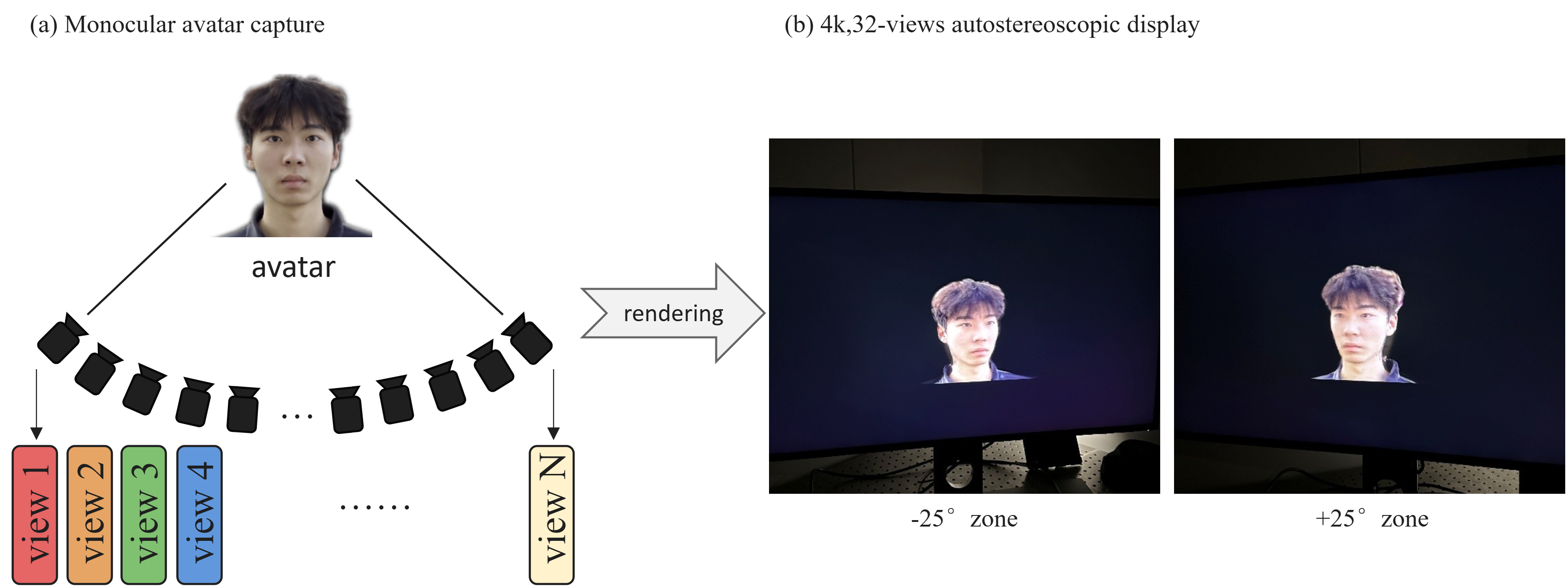}
\caption{\textbf{\method{} overview.}
(a) A monocular portrait sequence is used to reconstruct a Gaussian head avatar, which is rendered from multiple virtual display viewpoints.
(b) The rendered views are encoded into a 4K, 32-view glasses-free display raster; captured viewing zones at $-25^\circ$ and $+25^\circ$ show the view-dependent avatar appearance delivered by the lenticular panel.}
\label{fig:teaser}
\Description{Overview showing monocular avatar capture, virtual display views, rendering, and captured left and right viewing zones on a glasses-free 3D display.}
\end{teaserfigure}

\maketitle

\input{sec/1_intro}
\input{sec/2_related}
\input{sec/3_method}
\input{sec/4_experiments}
\input{sec/5_conclusion}

\bibliographystyle{ACM-Reference-Format}
\bibliography{main}

\end{document}

%% file: preamble.tex
\usepackage{graphicx}
\usepackage{stfloats}
\usepackage{placeins}

\newcommand{\method}{LentiAvatar}
\newcommand{\sideyaw}{\ensuremath{\pm 25^\circ}}
\definecolor{bestmetricshade}{RGB}{255,166,166}
\definecolor{secondmetricshade}{RGB}{255,213,169}
\newcommand{\bestscore}[1]{\begingroup\setlength{\fboxsep}{1pt}\colorbox{bestmetricshade}{\strut #1}\endgroup}
\newcommand{\secondscore}[1]{\begingroup\setlength{\fboxsep}{1pt}\colorbox{secondmetricshade}{\strut #1}\endgroup}

%% file: sec/0_abstract.tex
\begin{abstract}
Real-time stereoscopic video communication has long been a goal of immersive telepresence, yet practical systems still require specialized capture rigs or reduce remote users to a single portrait view. We present \method{}, a Gaussian head-avatar system that connects monocular avatar capture with subpixel-encoded glasses-free lenticular display for real-time autostereoscopic communication. From a monocular portrait video, \method{} reconstructs a controllable head avatar and optimizes it for the lateral viewing zones induced by the display. The method uses natural head turns as pseudo-multiview (PMV) supervision to constrain regions that are otherwise weakly observed in monocular training, including hair, ears, jaw contours, and neck boundaries. Reliable side frames are yaw-binned, aligned to virtual cameras, and supervised within a strict head-and-hair domain; contour-aware losses and staged regularization further suppress ghosting, alpha leakage, and depth instability while preserving lateral detail. At runtime, \method{} renders 32 virtual views and encodes them into a 4K lenticular raster with calibrated subpixel-routing masks. The live-tracker prototype sustains 10.65 FPS, and a subject-specific distilled driver raises the same display pipeline to 38.49 FPS.
\end{abstract}

%% file: sec/1_intro.tex
\section{Introduction}
\label{sec:intro}

Real-time portrait communication on glasses-free 3D displays requires the avatar pipeline to reconstruct, render, and route multiple views consistently. Earlier telepresence and portrait-reenactment systems connected real-time capture, tracking, and communication, but typically rely on specialized capture infrastructure or remain single-view video renderers~\cite{dou2016holoportation,thies2016face2face,kim2018deepvideo,thies2018headon}. Neural radiance fields and Gaussian head avatars have made personalized facial geometry and appearance practical from monocular video~\cite{mildenhall2020nerf,gafni2021dynamic,zheng2022imavatar,xiang2024flashavatar,qian2024gaussianavatars,feng2025gpavatar,li2025rgbavatar}. INSTA demonstrates minute-level monocular neural-head reconstruction~\cite{zielonka2023insta}, FlashAvatar and RGBAvatar study efficient monocular Gaussian avatars~\cite{xiang2024flashavatar,li2025rgbavatar}, and GaussianAvatars and GPAvatar demonstrate high-quality rigged or projection-efficient Gaussian heads~\cite{qian2024gaussianavatars,feng2025gpavatar}. On a lenticular display, however, the avatar is not judged from one frontal rendering. Multiple synthesized views are routed to calibrated raster subpixels and separated by optics into viewing zones, so side-view opacity, boundary, or color errors can appear as depth shimmer, ghosting, or color fringing.

This setting highlights a limitation in existing monocular avatar reconstruction. Front-dominant training video strongly constrains the face seen by the input camera, but it gives sparse evidence for the ears, hair silhouette, jawline, posterior head, and neck boundary. Recent Gaussian avatar systems render efficiently and with high frontal fidelity~\cite{xiang2024flashavatar,qian2024gaussianavatars,feng2025gpavatar,li2025rgbavatar}, yet their usual monocular objectives do not explicitly train the lateral views that a multiview display reveals. Directly adding side supervision can be unreliable because the head and hair approximately follow head rotation, while the neck, collar, and background do not; treating them alike can introduce neck-rear ghosting, alpha shells, skin-colored collar contamination, or blurred lateral contours.

Our key observation is that ordinary head turns already contain the missing side-view evidence when they are treated as view-specific supervision rather than as generic temporal frames. A turned frame observes the face, hair, ear, and jaw contour at a particular yaw, and can supervise the virtual camera with the same yaw after alignment refinement and above-neck masking. This converts monocular head motion into a practical pseudo-multiview (PMV) signal for the lateral views exposed by the display.

\begin{figure*}[t]
\centering
\includegraphics[width=\textwidth,trim=0 0bp 114bp 0,clip]{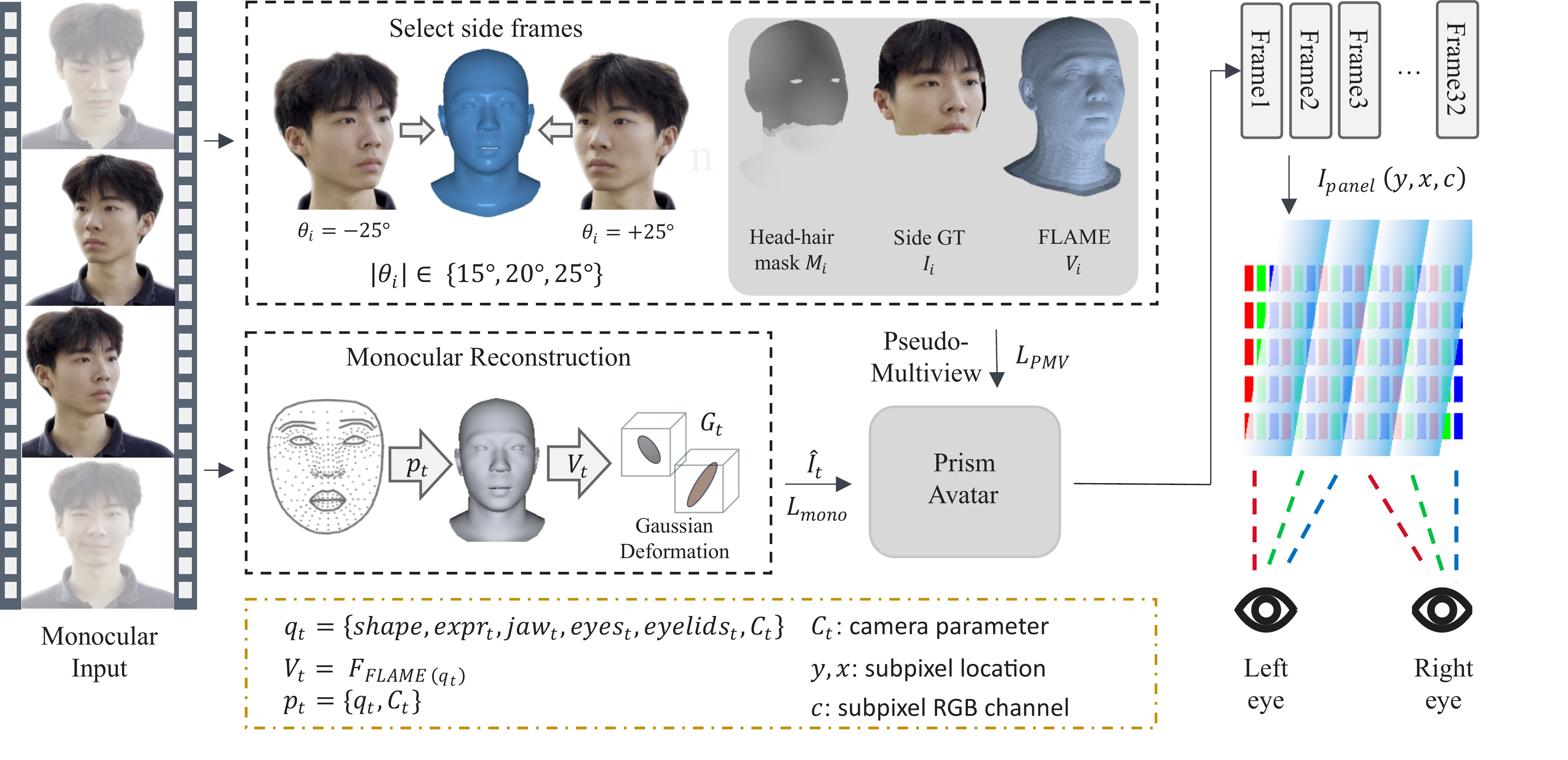}
\caption{\textbf{\method{} pipeline.}
Given a monocular portrait sequence, \method{} first reconstructs a controllable Gaussian head avatar from FLAME-based controls and monocular supervision.
Natural head turns are then selected as yaw-binned pseudo side observations, aligned to the corresponding virtual cameras, and restricted to an above-neck matte before applying the PMV loss.
At runtime, the trained avatar is driven by the current controls, rendered into multiple display views, and routed through calibrated subpixel masks to form the panel raster $I_{\mathrm{panel}}(y,x,c)$ for glasses-free stereoscopic presentation.}
\label{fig:pipeline}
\Description{System pipeline showing monocular input frames, side-frame selection, above-neck masking, pseudo-multiview supervision for a Gaussian head avatar, multiview rendering, subpixel mask encoding, and left/right-eye viewing on a glasses-free display.}
\end{figure*}

\method{} addresses stereoscopic video communication with a monocular Gaussian head avatar. During training, it selects natural head turns as PMV observations, ranks and snaps them to yaw bins, refines virtual-camera alignment, gates unreliable matches, and applies supervision inside a strict above-neck matte. Contour-preserving alpha and edge terms retain hair, ear, and jaw boundaries, while shell and collar controls suppress off-surface opacity and color leakage. During display-time stereoscopic rendering, the driven Gaussian avatar is rendered from virtual display views, encoded through calibrated subpixel masks, rasterized as a 4K panel frame, and presented as a glasses-free 3D avatar rather than a conventional single-view portrait. Figure~\ref{fig:pipeline} summarizes this reconstruction-to-display pipeline.

We evaluate \method{} with public benchmark comparisons, component analysis, and 4K display-side profiling. On Marcel, \method{} obtains the lowest outside-mesh alpha among the compared methods and ranks second on both neck-rear ghost measures, neck-rear smear, and alpha translucency smear. The live-tracking stereoscopic video communication system reaches 10.65 FPS in the 4K, 32-view configuration after initialization; a subject-specific student driver raises the same display pipeline to 38.49 FPS, confirming that live tracking is the dominant frame-time cost rather than subpixel composition.

Our contributions are:
\begin{itemize}
\item A reconstruction-to-display Gaussian head-avatar pipeline for glasses-free 3D communication that reconstructs a controllable avatar from monocular video and encodes 32 rendered views into a calibrated 4K autostereoscopic raster.
\item PMV supervision that turns natural monocular head rotations into yaw-matched side observations for lateral display views.
\item A reliability-aware side-view objective with strict head-and-hair masking, side-frame ranking, alignment gates, contour losses, and shell/collar regularization.
\end{itemize}

%% file: sec/2_related.tex
\section{Related Work}
\label{sec:related}

\subsection{Monocular and Gaussian avatars}
Parametric face models provide the tracking and semantic support used by many avatar systems, from 3D morphable models and FLAME to robust in-the-wild fitting~\cite{blanz1999morphable,li2017flame,feng2021deca}. Neural rendering and NeRF-based avatars learned dynamic face appearance from controlled or monocular capture~\cite{lombardi2018deepappearance,lombardi2019neuralvolumes,thies2019deferred,ma2021pixelcodec,mildenhall2020nerf,barron2022mipnerf360,muller2022instant,park2021nerfies,park2021hypernerf,gafni2021dynamic,zheng2022imavatar,athar2022rignerf,grassal2022nha,zheng2023pointavatar,duan2023bakedavatar,zielonka2023insta}, while real-time portrait reenactment and telepresence systems established the communication setting~\cite{dou2016holoportation,thies2016face2face,kim2018deepvideo,thies2018headon}. 3D Gaussian Splatting~\cite{zwicker2001surface,kerbl20233d} shifted avatar rendering toward explicit primitives, enabling deformable humans and head avatars with mesh embeddings, rigged Gaussians, efficient embeddings, parametric control, blendshape compression, relighting, tensorial appearance, or hybrid mesh-Gaussian editing~\cite{qian2024dgsavatar,hu2024hugs,shao2024splattingavatar,wang2025relightable,xiang2024flashavatar,chen2024monogaussianavatar,dhamo2024headgas,ma2024gaussianblendshapes,xu2024gaussianheadavatar,qian2024gaussianavatars,saito2024relightablegaussian,giebenhain2024npga,feng2025gpavatar,wang2025gaussianhead,moon2025geoavatar,li2025rgbavatar,zhang2025hravatar,wang2025tensorialgaussian,wang2025mega}. \method{} builds on this explicit-avatar line, but targets the display-coupled failure case where lateral viewing zones expose weakly constrained side contours and neck regions.

\subsection{Missing-view and pseudo-view priors}
Dense view coverage improves head reconstruction, as shown by multi-view datasets and systems such as NeRSemble~\cite{kirschstein2023nersemble}. Monocular and single-image methods instead infer unobserved regions with learned or generated priors, including diffusion-based Gaussian avatars, pseudo multi-view head synthesis, single-image Gaussian avatars, and related human or upper-body reconstruction methods~\cite{tang2025gaf,deng2024portrait4dv2,qiu2024anigs,li2025pshuman,zhuang2025idol,zhang2025guava}. These priors improve missing-view plausibility but are not tied to the physical viewing zones of a glasses-free panel. \method{} instead extracts real side evidence from natural head turns, then restricts it with yaw-binning, alignment gates, and a strict head-and-hair domain so that collar, lower-neck, and background pixels do not become side-view supervision.

\subsection{Glasses-free and subpixel display}
Glasses-free displays are rooted in light-field and image-based rendering, where a scene is sampled as directional views and routed to observer zones~\cite{levoy1996lightfield,gortler1996lumigraph,dodgson2005autostereoscopic,zwicker2006antialiasing}. Prior autostereoscopic and computational display systems studied dynamic-scene acquisition, multiview rendering, parallax barriers, multilayer/tensor displays, and tiled multi-view VR~\cite{matusik2004tv,jones2007rendering,kooima2010multiviewer,lanman2010contentadaptive,wetzstein2012tensor}. For lenticular panels, GPU interleaving, chroma or subpixel multiplexing, directional subpixel rendering, and recent light-field display rendering show that RGB subpixel layout, lens slant, and calibrated view routing are part of the rendering model~\cite{ruijters2008dynamic,marson2015chroma,pei2016subpixel,lee2018directional,yang2024directl}. Adjacent avatar-communication systems use dense capture or XR devices~\cite{dou2016holoportation,chen2025taoavatar}; our setting instead couples monocular avatar reconstruction, multiview Gaussian rendering, and calibrated subpixel routing for a live glasses-free panel.

%% file: sec/3_method.tex
\section{Method}
\label{sec:method}

The training input is a monocular portrait sequence $\{I_t,A_t,\mathbf{z}_t,y_t\}_{t=1}^{T}$, where $I_t$ is the RGB frame, $A_t$ is the foreground alpha or matte, $\mathbf{z}_t$ contains the tracked expression, jaw, eye, eyelid, and head-pose controls, and $y_t$ is the estimated camera-relative view yaw. We assume the monocular camera calibration is available from the base avatar pipeline and the glasses-free panel provides calibrated raster-routing constants. \method{} has two coupled goals: reconstruct side-stable avatar geometry and appearance from natural head turns, and encode the driven avatar into a calibrated subpixel raster for glasses-free binocular disparity.

The geometric assumption is conservative: the head and hair approximately co-rotate in selected side frames, while the lower neck, collar, and background are unreliable supervision domains. \method{} therefore uses temporal side observations only after yaw matching, alignment refinement, and strict above-neck masking.

\subsection{Gaussian avatar representation}
\paragraph{Gaussian splatting avatar}
We use an explicit Gaussian avatar because it supports fast multi-view rendering for the display path. Following surface splatting and 3D Gaussian Splatting~\cite{zwicker2001surface,kerbl20233d}, the avatar $\mathcal{G}=\{g_j\}_{j=1}^{M}$ stores anisotropic primitives with position $\boldsymbol{\mu}_j$, covariance $\boldsymbol{\Sigma}_j$, opacity $o_j$, and color features $\mathbf{c}_j$. As in 3DGS, the covariance is parameterized by a rotation $R_j$ and diagonal scale matrix $S_j$,
\begin{equation}
\boldsymbol{\Sigma}_j = R_j S_j S_j^{\top} R_j^{\top},
\label{eq:gaussian_covariance}
\end{equation}
which ensures a positive semidefinite anisotropic support for each primitive. The control code $\mathbf{z}_t$ deforms the canonical avatar into view-conditioned frame parameters,
\begin{equation}
\mathcal{G}_{t,\theta}
=\mathcal{D}_{\Theta}(\mathcal{G},\mathbf{z}_t,\theta),\qquad
(\hat{I}_{t,\theta},\hat{\alpha}_{t,\theta})
=\mathcal{R}(\mathcal{G}_{t,\theta},\Pi_{\theta}),
\label{eq:avatar_render_operator}
\end{equation}
where $\theta$ is the requested virtual-view yaw and $\Pi_{\theta}$ is the corresponding camera. We use the standard front-to-back alpha compositing of 3DGS and recent Gaussian avatar systems~\cite{kerbl20233d,xiang2024flashavatar,qian2024gaussianavatars,feng2025gpavatar,li2025rgbavatar}; the primitive representation is not the main novelty of \method{}. Our contribution is the display-coupled side-view supervision and routing. For side-view supervision, we project FLAME-derived semantic weights for lateral face, ears, chin, jawline, neck, and neck contours into pseudo-views, yielding reliable head support and high-risk neck/collar regions.

\subsection{Side-frame selection and pseudo-view cameras}
\paragraph{Yaw-binned selection}
\method{} constructs pseudo-view observations from natural head rotations rather than using every turned frame. Let $\mathcal{B}=\{15^\circ,20^\circ,25^\circ\}$ be the lateral bins used by our display probes. For each frame, we compute
\begin{equation}
\begin{aligned}
b_t&=\arg\min_{b\in\mathcal{B}}\left||y_t|-b\right|,\\
a_t&=\mathbf{1}\!\left[14^\circ\leq |y_t|\leq27^\circ\right]\,
\mathbf{1}\!\left[\left||y_t|-b_t\right|\leq2.5^\circ\right],
\end{aligned}
\label{eq:yaw_bin_selection}
\end{equation}
and accept frame $t$ only when $a_t=1$. The supervised virtual-view yaw and context rewrite are
\begin{equation}
\hat{y}_t=\operatorname{sign}(y_t)b_t,\qquad
\mathbf{z}_t^{\mathrm{pose}}\leftarrow(0,p_t),\quad
\mathbf{z}_t^{\mathrm{view}}\leftarrow(\hat{y}_t,0),
\label{eq:pseudo_view_context}
\end{equation}
where $p_t$ is the tracked pitch. Thus the frame provides a side-view observation for a yaw-matched virtual camera, while the avatar pose context remains near frontal. This keeps PMV supervision one-to-one and avoids treating a physical head turn as an ordinary monocular training view.

\paragraph{Side-frame ranking}
Yaw alone is not sufficient because large turns may contain motion, expression changes, poor matte boundaries, or collar contamination. We assign each yaw-valid candidate a score
\begin{equation}
q_t=\sum_m w_m\,Q_m(t)-\sum_n \lambda_n\,R_n(t),
\label{eq:side_frame_ranking}
\end{equation}
where the positive terms score yaw-bin agreement, temporal and expression stability, matte/edge cleanliness, and visible lateral face or hair support, while the risk terms penalize boundary instability and collar contamination. We keep only high-scoring frames per yaw bin and side direction, producing a compact, balanced PMV set with left and right coverage rather than trusting all side-looking frames equally.

\paragraph{Virtual cameras}
For both PMV supervision and runtime display, \method{} renders horizontal virtual views on a camera arc around the avatar. Let $\mathbf{c}_0$ be the avatar bounding-box center, $h$ the head-support height, and $\mathbf{e}_y$ the vertical axis. We use
\begin{equation}
\begin{aligned}
\mathbf{c}&=\mathbf{c}_0+0.08h\,\mathbf{e}_y,\\
\mathbf{o}(\theta)&=\mathbf{c}
+r[\sin\theta,0,\cos\theta]^{\top},\\
\Pi_{\theta}&=\operatorname{LookAt}(\mathbf{o}(\theta),\mathbf{c},K),
\end{aligned}
\label{eq:virtual_camera_arc}
\end{equation}
where $r$ is the calibrated avatar-camera radius and $K$ is the perspective intrinsic matrix. During training, $\theta=\hat{y}_t$; during display, the same construction samples 32 virtual views and routes them to the autostereoscopic raster in Sec.~\ref{sec:subpixel_display_encoding}.

\subsection{PMV reconstruction}
\paragraph{Camera and target alignment}
Even after yaw binning, real side frames and virtual cameras are not perfectly aligned because monocular tracking, cropping, and hair silhouettes are imperfect. \method{} therefore applies a bounded local search over camera yaw, radius, screen scale, and translation, followed by a small target-image similarity warp. Both steps maximize support overlap, boundary agreement, and centroid consistency while penalizing large corrections. The resulting side target is used only when the alignment gate passes:
\begin{equation}
g_t^{\mathrm{align}}=
\mathbf{1}\!\left[\operatorname{IoU}\geq0.32\right]\,
\mathbf{1}\!\left[\operatorname{IoU}_{e}\geq0.12\right]\,
\mathbf{1}\!\left[d_{\mu}\leq28\text{ px}\right],
\label{eq:alignment_gate}
\end{equation}
with additional implementation bounds on shift, scale, and rotation. Failed matches are downweighted rather than trusted as full supervision.

\paragraph{Strict head-and-hair supervision domain}
Let $\hat{I}_t$ and $\hat{\alpha}_t$ be the RGB image and alpha map rendered from the aligned pseudo-view camera, and let $I_t^{*}$ and $A_t^{*}$ be the aligned side-frame target. To exclude unreliable non-head regions, we first define the side-view support as a positive-minus-risk semantic mask:
\begin{equation}
P_t(u)=
\mathbf{1}\!\left[S_t(u)>\tau_s\right]\,
\mathbf{1}\!\left[R_t(u)<\tau_r\right],
\label{eq:side_support_mask}
\end{equation}
where $S_t$ rasterizes positive side-head weights and $R_t$ rasterizes neck and jaw-neck risk weights. The positive support covers the visible lateral face, ears, hair-adjacent head support, chin, and jawline; the risk support marks lower-neck, collar-adjacent, and jaw-neck regions that are unreliable in the monocular side frame. The strict PMV target matte is
\begin{equation}
M_t^{\mathrm{hh}}(u)=
A_t^{*}(u)\odot \mathcal{D}_{25}(P_t)(u)\odot B_t(u),
\label{eq:head_hair_matte}
\end{equation}
where $\mathcal{D}_{25}$ denotes a 25-pixel dilation and $B_t$ is a per-column bottom gate computed from the projected support boundary using a high support quantile, a small pixel extension, and a canonical FLAME height cutoff. This strict domain is the key difference from naive PMV: the RGB target may contain neck, collar, or background pixels, but those pixels do not supervise the side-view avatar.

\paragraph{Reliability-gated side objective}
Direct RGB side supervision is applied only in stable interior pixels. We erode the target matte, suppress boundary bands, partial-alpha pixels, and high-chroma regions, and remove RGB supervision from boundary pixels where small alignment errors could imprint color fringes on the Gaussian appearance. The scheduled training objective is
\begin{equation}
\begin{aligned}
\mathcal{L}(k)&=\mathcal{L}_{\mathrm{base}}+\gamma(k)\mathcal{L}_{\mathrm{pmv}},\\
\mathcal{L}_{\mathrm{pmv}}&=
\lambda_I\mathcal{L}_I+\lambda_{\alpha}\mathcal{L}_{\alpha}
+\lambda_{\mathrm{bg}}\mathcal{L}_{\mathrm{bg}}
+\lambda_{\mathrm{mesh}}\mathcal{L}_{\mathrm{mesh}}\\
&\quad+\lambda_e\mathcal{L}_e+\lambda_s\mathcal{L}_s .
\end{aligned}
\label{eq:training_objective}
\end{equation}
$\mathcal{L}_{\mathrm{base}}$ includes photometric, SSIM, foreground-alpha, background-alpha, and regularization terms. In $\mathcal{L}_{\mathrm{pmv}}$, $\mathcal{L}_I$ and $\mathcal{L}_{\alpha}$ supervise safe RGB and head-domain alpha pixels; $\mathcal{L}_{\mathrm{bg}}$ and $\mathcal{L}_{\mathrm{mesh}}$ suppress opacity outside the target support and dilated FLAME mesh; $\mathcal{L}_e$ preserves alpha gradients on ears, hair, and jaw boundaries; and $\mathcal{L}_s$ penalizes semi-transparent smear through the standard $4\hat{\alpha}(1-\hat{\alpha})$ response. For reproducibility, the base weights for L1, SSIM, foreground alpha, and background alpha are $(1.0,0.08,0.05,0.35)$. The PMV weights $(\lambda_I,\lambda_{\alpha},\lambda_{\mathrm{bg}},\lambda_{\mathrm{mesh}},\lambda_e,\lambda_s)$ are $(0.016,0.012,0.025,0.012,0.035,0.004)$ for side learning and $(0.016,0.004,0.012,0.006,0.026,0.010)$ for color fine-tuning. PMV starts at iteration 320, ramps linearly for 700 iterations, and is capped by $\gamma(k)\leq0.65$; the later no-PMV stages disable $\mathcal{L}_{\mathrm{pmv}}$. Side-aware shell/background cleanup is applied only on mesh and edge supports, using alpha strengths $0.05/0.07$ for the two PMV stages, color strength $0.012$, and boundary-chroma strength $0.85$; the collar anti-skin loss is disabled. This keeps the PMV objective localized to lateral artifacts rather than imposing broad penalties that would remove valid face, hair, ear, or jaw detail.

\paragraph{Staged stabilization}
\label{sec:staged_stabilization}
The final model uses a four-stage stabilization schedule: PMV side learning with auto-alignment, strict head-and-hair masking, shell cleanup, and contour preservation; PMV color fine-tuning with milder alpha/background weights; no-PMV low-weight monocular color correction with blend features and the weight module frozen; and a short no-PMV cleanup stage with base and blend color features frozen. This lets side-view evidence shape the avatar first, then removes residual color leakage without globally pruning the face, hair, ear, or jaw details learned from PMV supervision.

\subsection{Real-time driving and subpixel display encoding}
\label{sec:subpixel_display_encoding}
\paragraph{Subject-specific distilled driver}
The display prototype is limited by the per-frame metrical tracker, so we optionally distill the tracker into a subject-specific feed-forward RGB driver. The teacher targets are the 129-D grouped deformation controls already produced for the training sequence; a MobileNetV3-Small backbone with a two-layer MLP head predicts the normalized control vector from a cropped $224\times224$ portrait frame using a group-weighted smooth-$L_1$ distillation loss over expression, neck, jaw, eye, eyelid, and translation groups. At inference, the prediction is denormalized with subject statistics and passed to the same Gaussian deformation module as tracker output. The avatar, 32-view renderer, and subpixel compositor are unchanged, so this module only reduces the tracking-stage cost reported in Sec.~\ref{sec:experiments}.

\paragraph{Subpixel prism encoding}
We encode the multiview avatar output at the level of RGB subpixels.

\begin{figure}[t]
\centering
\includegraphics[width=\linewidth]{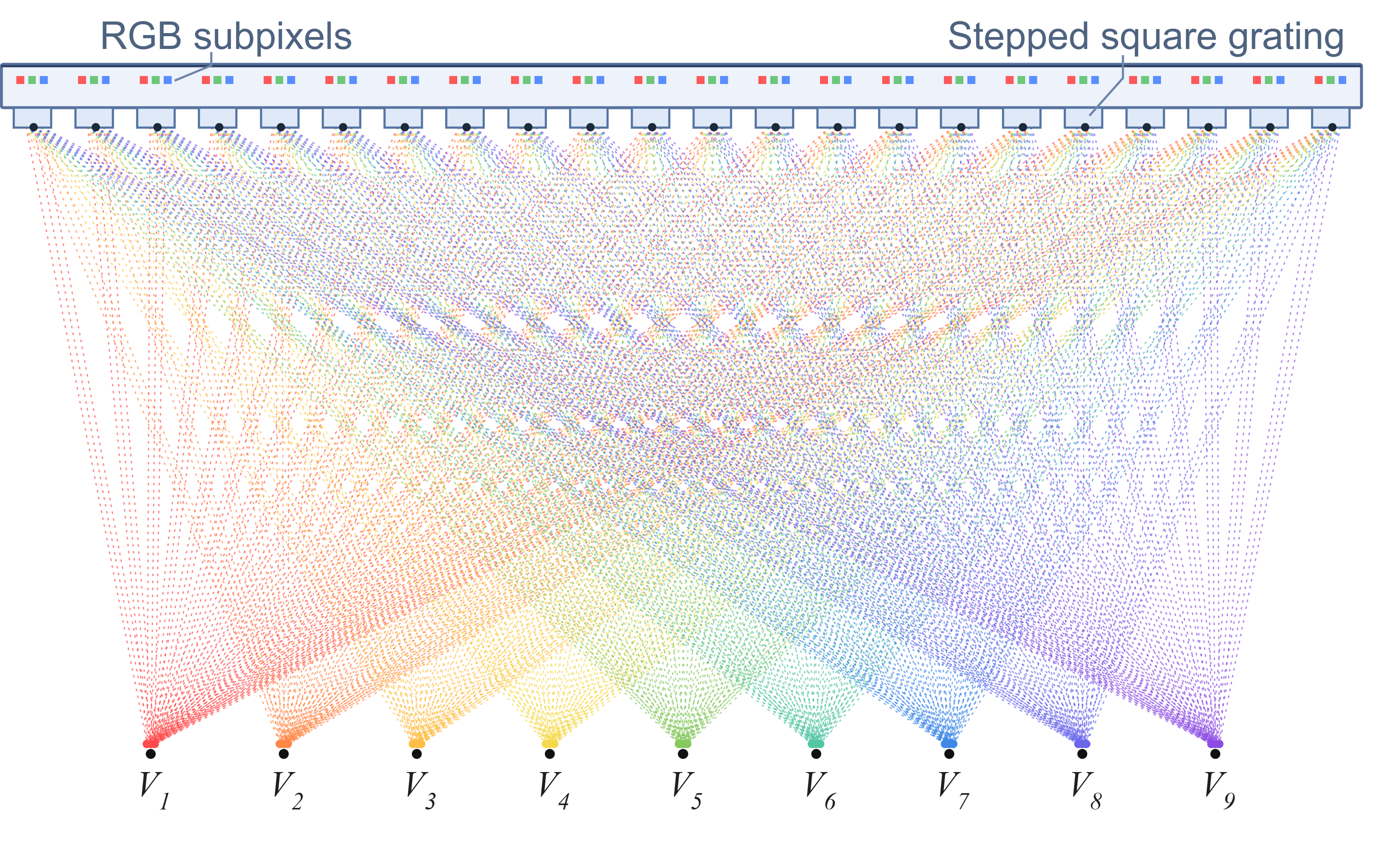}
\caption{\textbf{Subpixel-to-view routing in the glasses-free display.}
RGB subpixels lie underneath a calibrated grating layer, which redirects repeated subpixel positions toward different lateral viewing zones. Colored ray bundles indicate the angular distribution of the routed subpixels. \method{} uses this calibrated routing in the reverse direction: each rendered virtual avatar view is written only to the subpixels whose outgoing rays reach the corresponding viewing zone.}
\label{fig:multi_view_display}
\Description{A schematic of RGB subpixels under a stepped square grating. Colored rays from repeated subpixel positions are redirected toward multiple lateral viewing zones labeled V1 through V9.}
\end{figure}

Classic glasses-free displays create viewpoint-dependent images by optically directing different panel samples to lateral viewing zones~\cite{dodgson2005autostereoscopic,matusik2004tv,wetzstein2012tensor}. For lenticular, barrier, or grating-based panels, rasterization therefore becomes a view assignment problem at the subpixel level rather than conventional image compositing~\cite{pei2016subpixel,lee2018directional,yang2024directl}. Figure~\ref{fig:multi_view_display} illustrates the physical mapping: the grating couples subpixel position with outgoing angle, so neighboring RGB subpixels can contribute to different observer viewpoints. \method{} uses the calibrated panel constants to instantiate this mapping as binary routing masks for the virtual avatar views.

Let $E\in[0,1]^{H\times W\times 3}$ be the encoded output raster and let $V_i\in[0,1]^{H_v\times W_v\times 3}$ be the rendered image for virtual view $i\in\{1,\ldots,N\}$. A subpixel is denoted by $p=(x,y,c)$, where $(x,y)$ is the output pixel location and $c$ is the color channel after the configured RGB/BGR order. Let $\rho(c)\in\{0,1,2\}$ map the color channel to its subpixel offset inside an output pixel. The display constants are the raster slant coefficient $C_r$, the subpixel routing period $\Delta_r$, the reference offset $s_{\mathrm{ref}}$, and the reference view index $i_{\mathrm{ref}}$. For view $i$, the row-dependent offset is
\begin{equation}
s_i=s_{\mathrm{ref}}-\frac{\Delta_r}{N}(i-i_{\mathrm{ref}}).
\end{equation}
For row $y$ and integer period index $m$, the candidate subpixel coordinate is
\begin{equation}
b_i(y,m)=3C_r y+s_i+m\Delta_r.
\end{equation}
View $i$ contributes to subpixel $p$ when the subpixel index $x_s=3x+\rho(c)$ equals $\lfloor b_i(y,m)\rfloor$ for some $m$, and the fractional part of $b_i(y,m)$ is smaller than $\Delta_r/N$. This defines a binary mask $M_i(p)$. With $u_p$ the bilinear sample coordinate in the rendered view corresponding to output subpixel $p$, the encoded raster is
\begin{equation}
E(p)=\sum_{i=1}^{N} M_i(p)\,V_i(u_p).
\label{eq:display_encode}
\end{equation}

\paragraph{Runtime implementation}
The display implementation renders $N=32$ virtual views at $960\times540$, uniformly sampled over the avatar-viewing range $[-25^\circ,+25^\circ]$, and composes them into a $3840\times2160$ panel raster. The calibrated masks are precomputed once into sparse GPU sampling tables with primary and secondary view selectors; configurations with more than two active views at a subpixel are rejected. At each frame, the current controls deform the Gaussian avatar, the cached view renderers produce the virtual views, optional mesh-visibility gates suppress unreliable lateral background leakage, and alpha-boundary chroma attenuation reduces residual boundary color artifacts. Primary assignments use indexed overwrite into a half-precision output buffer, while secondary overlaps are accumulated by indexed addition. The final tensor is converted to the calibrated panel format for display.

%% file: sec/4_experiments.tex
\section{Experiments}
\label{sec:experiments}

\subsection{Setup}
The experiments test two claims: PMV reconstruction improves side-view robustness for monocular avatars, and the reconstructed avatar can be routed to a 4K glasses-free display at interactive rates. We compare qualitative behavior on public \textbf{INSTA} benchmark sequences~\cite{zielonka2023insta} with stable horizontal head turns against RGBAvatar~\cite{li2025rgbavatar}, HRAvatar~\cite{zhang2025hravatar}, FlashAvatar~\cite{xiang2024flashavatar}, and INSTA~\cite{zielonka2023insta}. These public-sequence comparisons are used to validate the reconstruction side of the method: if PMV supervision improves the side views reconstructed from monocular video, the same gains can be carried into the lateral viewing zones of the glasses-free display. Quantitative diagnostics use Marcel, rendered at $-25^\circ$ and $+25^\circ$ from frame 0 with a common aligned crop and 9-pixel FLAME-mask dilation.

For the component study, the avatar is trained with the four-stage schedule in Sec.~\ref{sec:staged_stabilization}: PMV side learning, PMV color fine-tuning, no-PMV low-weight color correction, and conservative no-PMV cleanup. The probe protocol follows the glasses-free display setting: $0^\circ$ checks the central viewing zone, while \sideyaw{} checks the lateral zones that are most sensitive to hair, ear, jaw, and neck artifacts.

\begin{figure*}[t]
\centering
\begin{minipage}{0.92\textwidth}
\centering
\includegraphics[width=\linewidth]{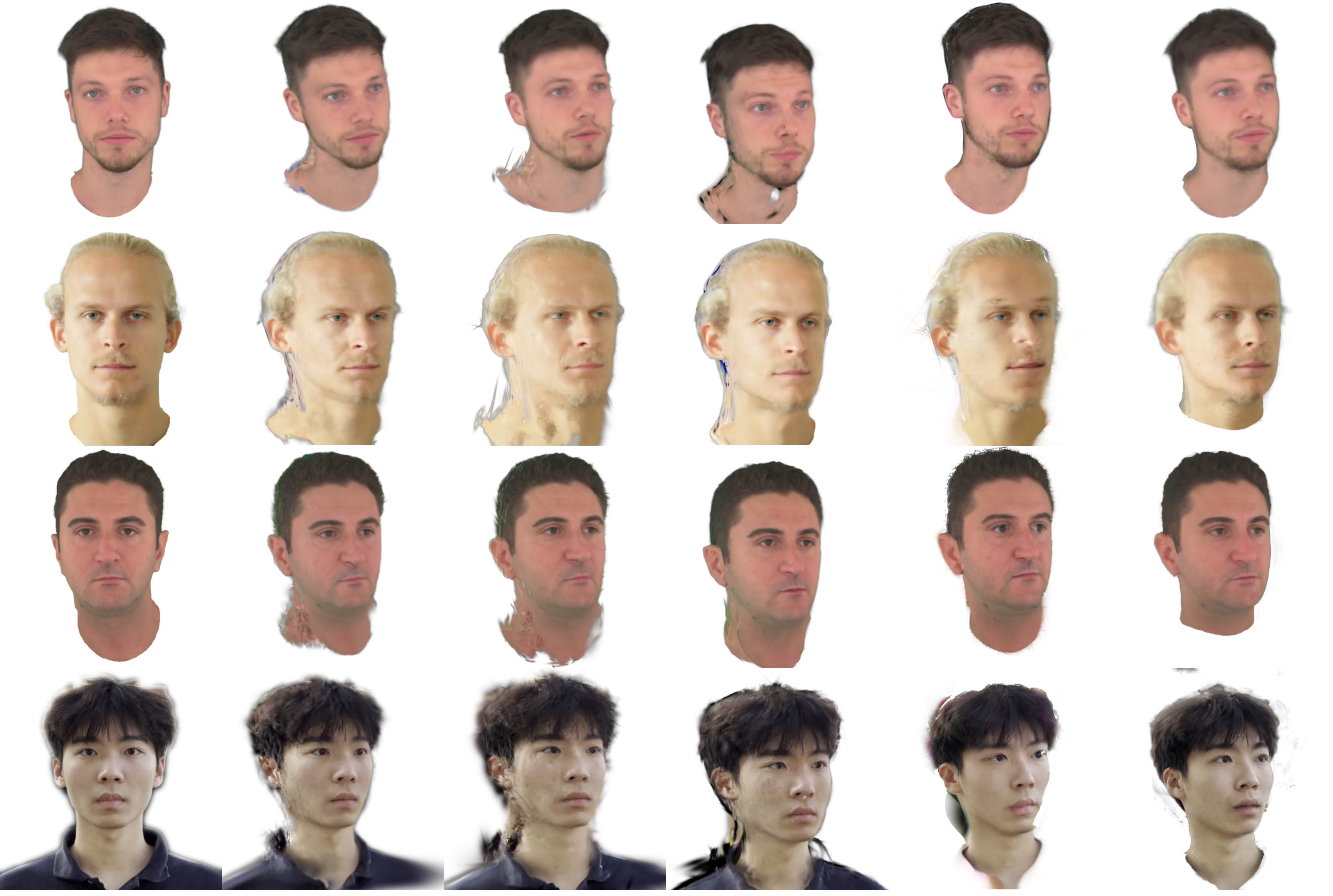}\\[-0.1em]
{\footnotesize
\makebox[\linewidth][c]{%
\makebox[0.166\linewidth][c]{GT}%
\makebox[0.166\linewidth][c]{Ours}%
\makebox[0.166\linewidth][c]{RGBAvatar}%
\makebox[0.166\linewidth][c]{HRAvatar}%
\makebox[0.166\linewidth][c]{FlashAvatar}%
\makebox[0.166\linewidth][c]{INSTA}}}
\end{minipage}
\caption{\textbf{Qualitative comparison on public benchmark sequences.}
Rows show representative subjects from the public \textbf{INSTA} benchmark sequences. The leftmost column shows the sequence reference image, and the remaining columns compare \method{} with recent monocular head-avatar baselines rendered at $-25^\circ$.
This lateral probe emphasizes the regime targeted by \method{}: preserving identity, hair, ear, and jaw structure while reducing neck leakage and boundary artifacts that become prominent in multiview autostereoscopic presentation.}
\label{fig:public_benchmark_comparison}
\Description{A multi-row qualitative benchmark comparison with columns for a reference image, our method, RGBAvatar, HRAvatar, FlashAvatar, and INSTA, showing side-view head-avatar renderings at negative 25 degrees.}
\end{figure*}

Figure~\ref{fig:public_benchmark_comparison} uses a fixed $-25^\circ$ lateral probe on smooth-turning INSTA sequences. The comparison shows that monocular baselines can preserve plausible reference-view appearance while producing floating opacity, weak side contours, or neck/collar contamination under large yaw. In our setting, this qualitative benchmark is not an endpoint by itself: it verifies that PMV reconstruction improves exactly the side-angle content that will later be replicated across the routed display views. These errors directly affect glasses-free presentation because neighboring synthesized views are routed to different viewing zones, where boundary leakage and side-view instability can appear as depth shimmer, ghosting, or color fringing during stereoscopic communication.

\subsection{Evaluation metrics}
\label{sec:evaluation_metrics}
Standard image metrics do not isolate the artifacts exposed by side viewing zones, so we report six lower-is-better diagnostics matched to Table~\ref{tab:marcel_artifacts}. For each side view, let $C$ be RGB color, $A$ be alpha, $M$ be the FLAME mesh mask, $D=\operatorname{dilate}(M,9)$, and $N=\{0.46\leq y/H\leq0.90,\,x/W\leq0.42\ \mathrm{or}\ x/W\geq0.58\}$. We measure outside support opacity, neck-rear ghosting, and translucent smear as
\small
\begin{equation}
\begin{aligned}
O_{\alpha}&=\frac{\sum_p A(p)(1-D(p))}{\sum_p A(p)},&
G_{\mathrm{m}}&=\frac{\sum_{p\in N}A(p)(1-D(p))}{\sum_p A(p)},\\
G_{\mathrm{d}}&=\frac{\sum_{p\in N}A(p)(1-D(p))}{|N|},&
S_{\mathrm{n}}&=\frac{\sum_{p\in N}4A(p)(1-A(p))}{|N|},\\
S_{\alpha}&=\frac{\sum_p 4A(p)(1-A(p))D(p)}{\sum_p D(p)} .
\end{aligned}
\end{equation}
\normalsize
$O_{\alpha}$ measures outside-support alpha; $G_{\mathrm{m}}$ and $G_{\mathrm{d}}$ measure neck-rear ghost mass and density; $S_{\mathrm{n}}$ and $S_{\alpha}$ measure translucency. Color heat uses saturation $s$, alpha boundary $B$ from $P(p)=\mathbf{1}[A(p)>0.03]$, and weighted mean $\operatorname{WM}(\cdot,\cdot)$:
\begin{equation}
\begin{aligned}
F_{\mathrm{c}}={}&0.45\,\operatorname{WM}(s,B A)
+0.35\,\operatorname{WM}(s,A(1-D))\\
&+0.20\,\operatorname{WM}(s,4A(1-A)D).
\end{aligned}
\end{equation}
For component analysis we additionally report side-contour edge energy $E_{\mathrm{c}}=\operatorname{WM}(\|\nabla C\|_1,BA)$, where higher values indicate sharper lateral boundaries.
Table~\ref{tab:marcel_artifacts} reports the Marcel mean over $-25^\circ$ and $+25^\circ$ renderings.

\begin{table}[H]
\caption{Marcel side-view artifact diagnostics. Values are averaged over $-25^\circ$ and $+25^\circ$ using a common crop and 9-pixel FLAME-mask dilation. \bestscore{Best} and \secondscore{second-best} results are highlighted.}
\label{tab:marcel_artifacts}
\centering
\small
\setlength{\tabcolsep}{1.15pt}
\renewcommand{\arraystretch}{1.03}
\begin{tabular*}{\columnwidth}{@{\extracolsep{\fill}}lcccccc@{}}
\toprule
Method & $O_{\alpha}\downarrow$ & $G_{\mathrm{m}}\downarrow$ & $G_{\mathrm{d}}\downarrow$ & $S_{\mathrm{n}}\downarrow$ & $S_{\alpha}\downarrow$ & $F_{\mathrm{c}}\downarrow$ \\
\midrule
\method{} & \bestscore{0.0691} & \secondscore{0.0237} & \secondscore{0.0297} & \secondscore{0.0304} & \secondscore{0.0149} & 0.2145 \\
RGBAvatar & 0.0768 & 0.0303 & 0.0384 & 0.0357 & 0.0182 & 0.2117 \\
HRAvatar & 0.0889 & 0.0314 & 0.0397 & \bestscore{0.0177} & \bestscore{0.0085} & \secondscore{0.2017} \\
FlashAvatar & \secondscore{0.0746} & 0.0368 & 0.0435 & 0.0851 & 0.0901 & \bestscore{0.1685} \\
INSTA & 0.0779 & \bestscore{0.0227} & \bestscore{0.0280} & 0.0330 & 0.0270 & 0.2315 \\
\bottomrule
\end{tabular*}
\end{table}

Table~\ref{tab:marcel_artifacts} compares side-view reconstructions on Marcel. The most direct display-facing signal is outside-support opacity: floating alpha outside the mesh is routed to neighboring viewing zones and can appear as a halo or ghost layer on the lenticular panel. \method{} gives the lowest $O_{\alpha}$ and remains competitive on both neck-rear ghost measures, which is consistent with the strict head-and-hair supervision domain. INSTA attains slightly lower neck-rear ghost mass and density, and HRAvatar or FlashAvatar are lower on selected translucency or color-fringe scores, but these gains do not uniformly translate to cleaner lateral support in Fig.~\ref{fig:public_benchmark_comparison}. The result is therefore best read as a display-specific trade-off rather than a generic image-metric leaderboard: \method{} suppresses off-surface side opacity while preserving enough lateral structure for the 32-view routing pipeline. This is the aspect that matters most to stereoscopic video chat, because each conversational frame is expanded into many laterally routed views; artifacts that remain near the neck boundary, collar, or outer silhouette are therefore repeatedly exposed across viewing zones instead of staying hidden in a single frontal portrait.

\subsection{Component-wise construction of PMV reconstruction}
\begin{figure*}[t]
\centering
\begin{minipage}{0.92\textwidth}
\centering
\includegraphics[width=\linewidth]{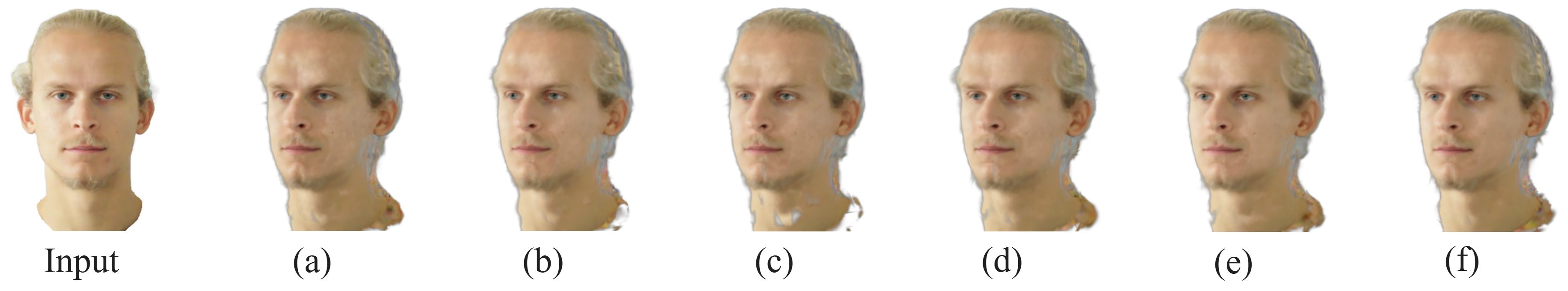}
\end{minipage}
\caption{\textbf{Progressive PMV ablation.}
The input is followed by lateral-view renderings from (a) no PMV, (b) naive PMV, (c) strict head-and-hair matte supervision, (d) side-frame ranking with yaw-bin selection, (e) alignment and camera refinement, and (f) the final \method{} model with staged stabilization.}
\label{fig:component_ablation}
\Description{A component ablation montage showing input, no pseudo-multiview supervision, naive pseudo-multiview supervision, head-and-hair masking, side-frame ranking and yaw-bin selection, alignment refinement, and the final LentiAvatar result.}
\end{figure*}

Figure~\ref{fig:component_ablation} and Table~\ref{tab:component_ablation} isolate the contribution of each reconstruction component. The no-PMV baseline lacks lateral evidence and therefore under-constrains side contours at \sideyaw{}. Naive PMV introduces yaw-matched side supervision and already increases side-contour energy from 0.1201 to 0.1234, but it also raises outside-support alpha from 0.0625 to 0.0649 and color-fringe heat from 0.2412 to 0.2474, showing that side views alone are not yet safe supervision. Adding the strict head-and-hair matte yields the cleanest intermediate neck-rear ghost scores by removing lower-neck and collar regions from the pseudo-view target, while frame ranking and yaw-bin selection further regularize which natural turns are allowed to supervise each lateral view. Alignment refinement then reduces residual real-to-virtual mismatch, especially in the lateral boundary region where small offsets would otherwise imprint edge ghosts or color pull.

The final staged model combines these components with the later no-PMV cleanup schedule to achieve the best overall artifact-detail balance. Relative to No PMV, \method{} reduces outside-support alpha by 2.9\%, alpha translucency smear by 3.1\%, and color-fringe heat by 4.6\%, while increasing side-contour energy by 5.6\%. This progression is consistent with the qualitative trend in Fig.~\ref{fig:component_ablation}: the early PMV stages recover missing side structure, and the later stabilization stages retain that structure while removing residual opacity leakage and chromatic boundary artifacts.

\begin{table}[H]
\caption{Component ablation on Marcel at \sideyaw{}. Metrics follow Sec.~\ref{sec:evaluation_metrics}; all are lower-is-better except $E_{\mathrm{c}}$.}
\label{tab:component_ablation}
\centering
\small
\setlength{\tabcolsep}{0.6pt}
\renewcommand{\arraystretch}{1.02}
\begin{tabular*}{\columnwidth}{@{\extracolsep{\fill}}lccccccc@{}}
\toprule
Variant & $O_{\alpha}\downarrow$ & $G_{\mathrm{m}}\downarrow$ & $G_{\mathrm{d}}\downarrow$ & $S_{\mathrm{n}}\downarrow$ & $S_{\alpha}\downarrow$ & $F_{\mathrm{c}}\downarrow$ & $E_{\mathrm{c}}\uparrow$ \\
\midrule
No PMV & \secondscore{0.0625} & 0.0170 & 0.0223 & 0.0276 & \secondscore{0.0187} & 0.2412 & 0.1201 \\
Naive PMV & 0.0649 & 0.0167 & 0.0215 & \bestscore{0.0255} & 0.0239 & 0.2474 & 0.1234 \\
+HH matte & 0.0631 & \bestscore{0.0161} & \bestscore{0.0204} & 0.0364 & 0.0409 & 0.2403 & \secondscore{0.1242} \\
+Rank/bin & 0.0646 & 0.0172 & 0.0225 & \secondscore{0.0263} & 0.0202 & 0.2477 & 0.1233 \\
+Align & 0.0630 & 0.0186 & 0.0242 & 0.0269 & 0.0195 & \secondscore{0.2374} & 0.1233 \\
\method{} & \bestscore{0.0607} & \secondscore{0.0164} & \secondscore{0.0214} & 0.0291 & \bestscore{0.0181} & \bestscore{0.2300} & \bestscore{0.1268} \\
\bottomrule
\end{tabular*}
\end{table}

\FloatBarrier

\subsection{Autostereoscopic display profiling}
For each control frame, the runtime renderer generates 32 avatar views at $960\times540$ and subpixel-composes them into a $3840\times2160$ raster. On an RTX 4090 after initialization, Table~\ref{tab:runtime_profile} compares live metrical tracking with the subject-specific RGB student driver from Sec.~\ref{sec:subpixel_display_encoding}. The comparison uses the same trained avatar, view set, renderer, and subpixel compositor, isolating the cost of estimating the driving controls from the cost of the display pipeline.

\begin{table}[H]
\caption{Runtime profile. Timings are in milliseconds except FPS; stage timings are independently measured counters rather than an additive budget.}
\label{tab:runtime_profile}
\centering
\small
\setlength{\tabcolsep}{2.8pt}
\renewcommand{\arraystretch}{1.03}
\begin{tabular*}{\columnwidth}{@{\extracolsep{\fill}}lcc@{}}
\toprule
Runtime component or metric & Live tracker & Student driver \\
\midrule
End-to-end FPS $\uparrow$ & 10.65 & 38.49 \\
Tracking stage $\downarrow$ & 109.15 & 4.68 \\
Render 32 views $\downarrow$ & 18.59 & 16.09 \\
Subpixel composition $\downarrow$ & 1.67 & 1.67 \\
Display presentation $\downarrow$ & 1.23 & 0.79 \\
Total loop $\downarrow$ & 109.15 & 26.15 \\
\bottomrule
\end{tabular*}
\end{table}

The profile shows that the live prototype is not bottlenecked by subpixel routing. Composing the 4K lenticular raster takes only 1.67 ms, and rendering all 32 views remains below 19 ms in both settings. Instead, the live metrical tracker dominates the frame time: its 109.15 ms stage cost limits the full system to 10.65 FPS even after initialization. Replacing this tracker with the subject-specific student driver reduces the driving stage to 4.68 ms and raises the same 4K, 32-view display pipeline to 38.49 FPS. For stereoscopic video chat, this difference is practically important because conversational quality depends on how smoothly the avatar controls can be refreshed, not only on how quickly the panel image can be composed. The student-driver setting therefore shows that once the driving cost is reduced, the same avatar renderer and calibrated subpixel encoder can support a much smoother live communication prototype; future system work should primarily reduce or generalize the control-estimation stage.

%% file: sec/5_conclusion.tex
\section{Limitations and Future Work}
\label{sec:limitations}

\method{} relies on informative horizontal head turns in the monocular training sequence. If side frames are sparse, misaligned, or affected by expression changes, hair motion, or poor mattes, PMV supervision weakens. The strict above-neck domain reduces collar and lower-neck contamination, but also limits lower-neck and clothing reconstruction; side-view quality remains below synchronized multi-view capture, especially for challenging hair, ears, and rear-neck regions.

The real-time prototype is also constrained by avatar driving. Live metrical tracking dominates frame time, while the distilled driver improves throughput but still requires subject-specific samples and a separate distillation stage. Future work will reduce tracking and multiview rendering latency, improve perceived realism on the glasses-free panel, jointly tune virtual-camera baselines with panel routing, and extend the head-only setting toward full-body stereoscopic communication.

\section{Conclusion}
\label{sec:conclusion}

We presented \method{}, a monocular Gaussian head-avatar system for subpixel-routed glasses-free stereoscopic communication. It converts natural head turns into strict above-neck PMV supervision, reducing lateral ghosting and alpha leakage while preserving side detail. At runtime, the trained avatar is encoded into a 4K autostereoscopic raster, and a subject-specific distilled driver supports real-time 32-view display above 30 FPS.